\documentclass[sigconf, anonymous=False]{acmart}
\usepackage{graphicx}
\usepackage{latexsym}
\usepackage{paralist}
\usepackage{url}
\usepackage{subcaption}
\usepackage{amsthm}
\usepackage{soul}

\theoremstyle{definition}
\newtheorem*{dfn*}{Definition}

\AtBeginDocument{%
  \providecommand\BibTeX{{%
    \normalfont B\kern-0.5em{\scshape i\kern-0.25em b}\kern-0.8em\TeX}}}

\begin{document}

\title{Area Modeling using Stay Information for Large-Scale Users and Analysis for Influence of COVID-19}

% \affiliation{%

\author{Kazuyuki Shoji}
\email{shoji@ucl.nuee.nagoya-u.ac.jp}
\affiliation{%
  \institution{Nagoya University}
  \country{Japan}
}

\author{Shunsuke Aoki}
\email{shunsuke@nagoya-u.jp}
\affiliation{%
  \institution{Nagoya University}
  \country{Japan}
}

\author{Takuro Yonezawa}
\email{takuro@nagoya-u.jp}
\affiliation{%
  \institution{Nagoya University}
  \country{Japan}
}

\author{Nobuo Kawaguchi}
\email{kawaguti@nagoya-u.jp}
\affiliation{%
  \institution{Nagoya University}
  \country{Japan}
}


\begin{abstract}
  Understanding how people use area in a city can be a valuable information in a wide range of fields, from marketing to urban planning.
Area usage is subject to change over time due to various events including seasonal shifts and pandemics.
Before the spread of smartphones, this data had been collected through questionnaire survey.
However, this is not a sustainable approach in terms of time to results and cost.
There are many existing studies on area modeling, which characterize an area with some kind of information, using Point of Interest (POI) or inter-area movement data.
However, since POI is data that is statically tied to space, and inter-area movement data ignores the behavior of people within an area, existing methods are not sufficient in terms of capturing area usage changes.
In this paper, we propose a novel area modeling method named Area2Vec, inspired by Word2Vec, which models areas based on people's location data.
This method is based on the discovery that it is possible to characterize an area based on its usage by using people's stay information in the area.
And it is a novel method that can reflect the dynamically changing people's behavior in an area in the modeling results.
We validated Area2vec by performing a functional classification of areas in a district of Japan.
The results show that Area2Vec can be usable in general area analysis.
We also investigated area usage changes due to COVID-19 in two districts in Japan.
We could find that COVID-19 made people refrain from unnecessary going out, such as visiting entertainment areas.

% https://www.editage.jp/insights/the-secret-to-using-tenses-in-scientific-writing
% https://blog.wordvice.jp/which-tense-should-be-used-in-abstracts-past-or-present/
% https://www.gfd-dennou.org/member/hiroki/homepage-old2/main027.html
\end{abstract}





\maketitle
\pagestyle{empty}

\section*{About This Paper}
This paper is an English translation of the paper published in the Transactions of the Information Processing Society of Japan (http://doi.org/10.20729/00213190).
\\

% 1
\section{Introduction} \label{sec:intro}
A city can be considered an aggregation of areas with various usage, such as office, residential, commercial, and entertainment areas.
However, the usage of these areas changes over time due to various factors like urban development, seasonal changes, and pandemics.
COVID-19 is a well-known example.
During the pandemic, people refrained from going out, and the number of people from restaurants and entertainment establishments decreased.
Additionally, the rise of remote work environments led to people spending more time at home throughout the day and night.
Analyzing area usage can detect such changes, providing valuable information for marketing strategies, urban planning, and government policies.
In area modeling, it's necessary to accurately reflect the functions of different areas, such as office or shopping areas, and capture changes in their usage over time.
In addition, it is desirable to represent areas as vectors since the results of area modeling are expected to be used for calculations such as similarity calculations and to be converted to machine learning features.

In existing research on area modeling, Point of Interest (POI) data~\cite{Yuan2012DiscoveringRODF,ZHAI2019BeyondW2V,Yao2016SensingSD} and inter-area transition data~\cite{Pan2013LanduseCUT,Yao2018RepresentingUF,Crivellari2019FromMA} have been extensively used. 
However, POI data has the drawback that it can not capture area usage changes due to shifts in people's behavior because it is statically defined information tied to space.
Moreover, areas without POIs, such as residential zones, are often excluded from the modeling process. 
Regarding inter-area transition data, it only provides information about movements from one area to another. 
This results in a focus solely on the strength of connections between areas, without considering the activities of people within each area.

\begin{figure}[t]
    \includegraphics[width=\linewidth]{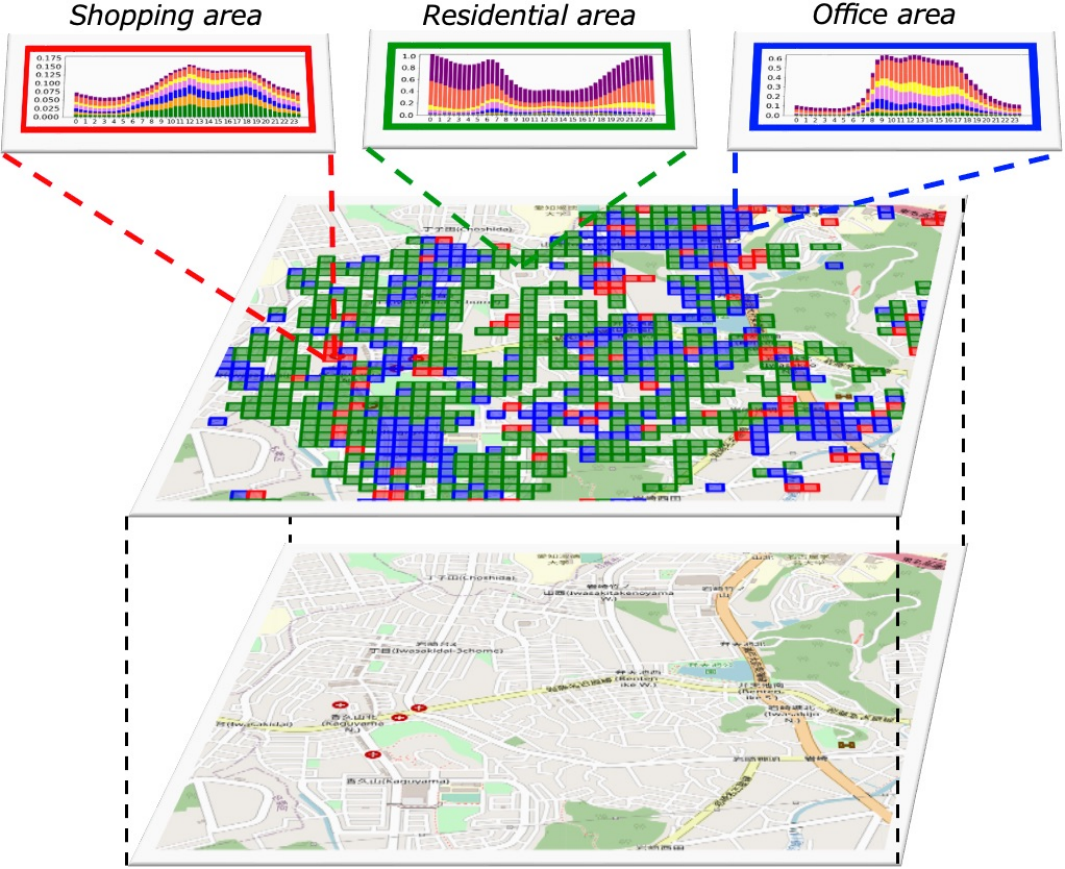}
    \caption{The overview for our area modeling. This includes meshing the target area, creating area embeddings using stay temporal information, clustering, graphing each cluster, and providing interpretations.}
    \label{fig:process_for_areamodeling}
\end{figure}

To address this challenge, we propose Area2vec, a model that uses stay information to characterize areas by its usage.
Area2Vec, inspired by the Word2Vec~\cite{mikolov2013distributed}, creates distributed representations of areas.
We can presume that areas where people spend long hours from morning to evening on weekdays are office areas.
We can also presume that areas where people spend long hours in the night regardless of whether weekdays or weekends are residential areas.
The model is based on this fact, and is designed so that areas with similar usage are placed close to each other in vector space.
The stay information includes the day of the week, arrival time, and stay time.
This approach enables the characterization of areas in response to various temporal events without relying on data other than location data.
Figure \ref{fig:process_for_areamodeling} outlines our area modeling process.
Initially, areas are characterized by their usage using the stay information and a distributed representation of areas is created.
In this paper, an area distributed representation embedded its usage from the stay perspective is termed UAS (Usage of Area with Stay information).
However, UAS, in its raw form, is a mere sequence of numbers without interpretable meaning.
Thus, the next step involves clustering UASs to abstract the information. 
Finally, a specialized graphing is performed to allow for the assignment of interpretations of what features each area has.
For example, UASs assigned to Cluster \#1 may be interpreted as "residential areas," while UASs assigned to Cluster \#2 may be interpreted as "office areas."

We conducted an experiment to evaluate Area2Vec from the following two perspectives:
\begin{inparaenum}[(i)]
    \item Is it possible to capture the characteristics of the function of each area? \label{process_1}
    \item Is it possible to capture area usage changes due to the effects of events over time? \label{process_2}
\end{inparaenum}
The location data we used was GPS data collected from authorized users' smartphones, focusing on experiments in Kokubuncho, Sendai, Miyagi, and Kabukicho, Shinjuku, Tokyo, Japan.
Regarding (\ref{process_1}), we confirmed that it is possible to classify areas for each function consistently with human senses.
This result demonstrates that area modeling method can be achieved from only location data.
As for (\ref{process_2}), we confirmed that it is possible to capture area usage changes due to the influence of COVID-19, and furthermore, to explain what kind of changes these are.
This result demonstrates the effectiveness of Area2Vec in fields like marketing and government policy, where rapid detection of situational changes is essential.

The main contributions are summarized as follows.
\begin{enumerate}
    \item We propose a novel area modeling method Area2Vec, which characterizes areas based on their usage and embeds their characterizations into vectors by utilizing people's stay information.
    \item We validate the usable of Area2Vec through the use of large location data to classify urban functions.
    \item We demonstrate the utility of Area2Vec by confirming that it can analyze changes in people's behavior by observing area usage changes due to the impact of COVID-19.
\end{enumerate}

The rest of this paper is organized as follows.
We review the related works in Section 2, and explain our area modeling method in Section 3.
In Section 4, we describe about dataset and conduct an evaluation experiment of Area2Vec. 
We finally provide a summary and discuss future prospects in Section 5.

\begin{figure*}[t]
    \begin{center}
      \includegraphics[width=\linewidth]{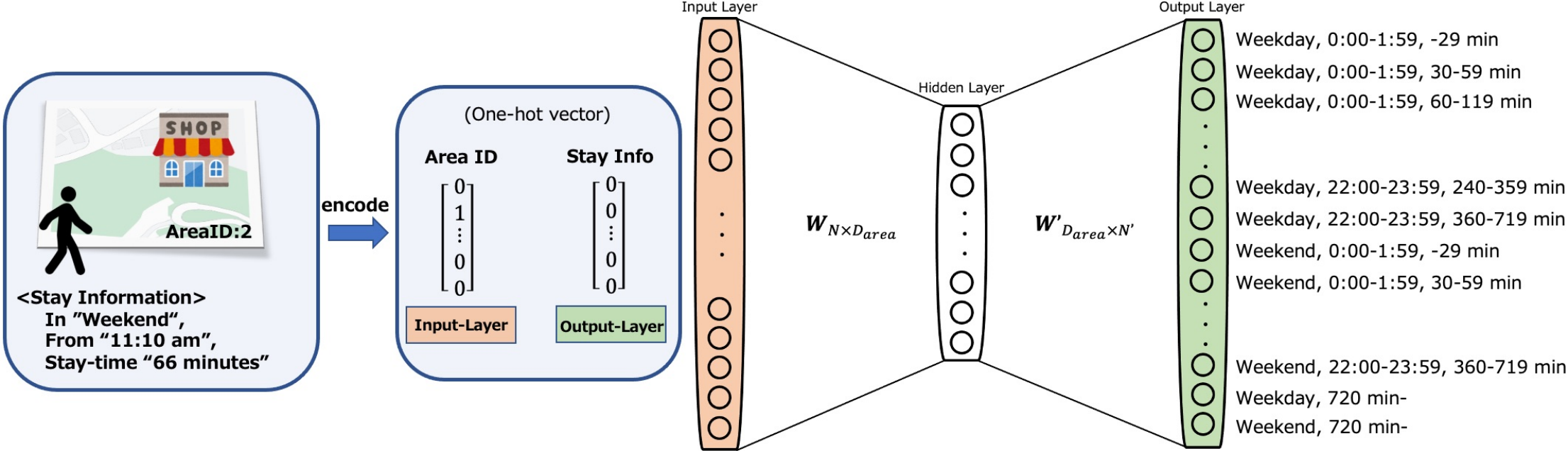}
      \caption{The architecture of Area2Vec. Area2Vec is inspired by the Word2Vec algorithm. During the training phase, when a area is input, the model predicts the stay temporal information for that area. After the training is completed, the \textit{W} component becomes area embeddings.}
      \label{fig:Area2Vec}
    \end{center}
\end{figure*}

% 2
\section{Related Work} \label{sec:relatedwork}
Human location data have been collected from diverse sources, leading to active research related to urban and human mobility~\cite{Toch2018AnalyzingLH,Zhou2018UnderstandUHM,Wang2019UHM}. 
For instance, user modeling~\cite{Becker2013HumanMC,Esuli2018Traj2UserEE, Xiao2010FindingSU, Zion2018IdentifyingAP}, estimating potential customers~\cite{Feng2017POI2VecGL}, predicting urban dynamics~\cite{Shimosaka2015ForecastUD}, POI recommendation~\cite{Li2016POIRecommend}, and anomaly detection in urban areas~\cite{Zhang2018DetectionUA} are among the many applications.
The demand for leveraging location data extends across urban planning, marketing, traffic management, disaster preparedness, and more. 

Our focus lies in area modeling, aiming to characterize each area in a city from some perspective.
Yuan et al.~\cite{Yuan2012DiscoveringRODF} proposed the DRoF (Discovers Regions of different Functions) method, utilizing taxi boarding and alighting data along with POI data and employing topic models to discover regions with distinct functionalities.
Pan et al.~\cite{Pan2013LanduseCUT}, criticizing the limitations of traditional urban function classification using remote sensing data, such as a limited number of classification categories and the inability to reflect human behaviors, conducted urban function classification using taxi boarding and alighting data.
Yao et al.~\cite{Yao2018RepresentingUF} and Crivellari et al.~\cite{Crivellari2019FromMA} proposed a method using inter-area transition data and applying Word2Vec to consider the before-and-after relationships of area transitions, creating a distributed representation of areas.
Zhai et al.~\cite{ZHAI2019BeyondW2V} employed Place2Vec~\cite{Yan2017FromITDL} to create a distributed representation of POIs, classifying urban functions at the scale of neighborhood area.

The aforementioned studies can be categorized into two groups; those that model the functions of each area in a city based on POI data and those that model the strength of connections and proximity of roles between areas based on inter-area transition data.
However, our work differs from conventional area modeling methods by characterizing areas based on people's stay information and modeling them according to their usage.
Using stay information provides a area modeling method that can detect changes in people's behavior over time due to various events, unlike approaches using less responsive subjects to external factors like POI data.
Additionally, inter-area transition data often relies on taxi boarding and alighting data.
This data may have a drawback as the place where a person alights is not necessarily the same as their intended destination.
Moreover, using only the fragmented information from the moment of boarding and alighting can result in a lack of knowledge about the behaviors taken before boarding or after alighting.
Consequently, when performing area modeling, there is a potential for inaccuracies in reflecting features.
In contrast, our proposed method Area2Vec utilizes stay information, which closely resembles area usage in terms of human behaviors, allowing for detailed and accurate area characterization.

% 3
\section{Area2Vec} \label{sec:methodology}
\begin{table}[t]
    \caption{Stay information used in our area modeling.}
    \begin{tabular}{|p{1.7cm}|p{6cm}|} \hline
        Day of Week & Weekday, Weekend (including holidays) \\ \hline
        Arrival Time & 0:00-1:59, 2:00-3:59, …, 20:00-21:59, 22:00-23:59 \\ \hline
        Stay Time & -29 min, 30-59 min, 60-119 min, 120-240 min, 240-359 min, 360-719 min, 720 min- \\ \hline
    \end{tabular}
    \label{table:time_info_on_stay}
\end{table}

We describe the details of Area2Vec in this section.
Area2Vec is inspired by the Word2Vec.
Word2Vec is a method for creating a distributed representation of words by solving the task of predicting which words come around a target word, based on the hypothesis that the surrounding words of the target word hold significant semantics about the target word.
Utilizing this technique, methods have been proposed to make distributed representations of various domains (users~\cite{Esuli2018Traj2UserEE}, POIs~\cite{Yan2017FromITDL}, regions~\cite{Crivellari2019FromMA}).
Area2Vec aims to create a distributed representation of areas based on the idea that people's stay information is influential in characterizing the area usage.
Stay information utilized in this paper includes the day of the week, arrival time, and stay time.
While Word2Vec includes skip-gram and continuous-bag-of-words, we describe Area2Vec under the assumption of using skip-gram.

This paper aims to create a distributed representation of areas that reflect the characteristics of area usage using stay information.
However, there are constraints to consider.
One limitation is that when there are two areas with facilities belonging to the same category, but with different people's stay patterns, these two areas may be distantly positioned in the vector space.
For example, if there are two restaurants, where one allows entry without queuing and the other often has long waiting times due to queues, their representations would not be positioned closely in the vector space. 
Another limitation is that it does not take into account the influence of interactions between areas.
For example, many people commute between their homes and offices on a daily basis, and it is commonly believed that there is a relationship between these two areas from a connectivity perspective.
However, since our method relies solely on stay information for areas, it cannot consider this relationship.
Therefore, by focusing on the area modeling based on features aligned with the actual patterns of facility usage, this paper is positioned as complementary to the methods having the above shortcomings.

The architecture of Area2Vec is showed in Figure \ref{fig:Area2Vec}.
The weights between the input and hidden layers and between hidden and output layers are represented as a matrix of shape $\textit{N}\times\textit{D}_{area}$ and  $\textit{D}_{area}\times\textit{N'}$, respectively, where \textit{N}, \textit{N'} and $\textit{D}_{area}$ denote the number of areas to be modeled, the number of nodes in the output layer, and the dimension of area vector, respectively.
The number of \textit{N'} varies depending on the information used.
This paper uses the temporal information related to stay shown in Table \ref{table:time_info_on_stay}.
Hence, each node in the output layer has the following unique time information: "\textit{stay for $\beta$ hours from $\alpha$ o'clock on weekdays (or weekends)}."
However, in cases where the stay time exceeds 720 minutes, an exception is made by disregarding the arrival time and only considering the day of the week.
The reason is that the number of stays exceeding 720 minutes is small, and if arrival time is taken into account in this situation, few predictions targeting the same stay information are generated during training, and appropriate correlations may not be obtained between areas.
Under this strategy, when an area is input, the parameters are trained to predict stay information in that area.
After training, the weights between the input and hidden layers, i.e., \textit{W}, become area embeddings that reflect how people use the areas.
Here, a term is defined.
\begin{dfn*}
  UAS (Usage of Area with Stay information): This refers to embeddings of each area generated from Area2Vec.
\end{dfn*}
\noindent
It is named as such because an area embedding reflects the area usage from the perspective of stay information.
The dimension of UAS is a hyperparameter, set to 4 in this paper.
This choice is based on the formula mentioned in \cite{googleDevelopers}:
$$\mbox{$embedding\_dimensions$}=\mbox{$number\_of\_categories$}^{0.25}$$

According to this formula, the recommended dimension of embeddings is the fourth root of the number of categories.
In Word2Vec, the number of categories corresponds to the number of words, but in Area2Vec, considering all combinations of information shown in the Table \ref{table:time_info_on_stay}, which corresponds to \textit{N'} and is 146, applying the formula results in 3.48.
However, since it must be an integer and rounding down to 3 may not capture all the information, a minimum of 4 dimensions is deemed necessary to adequately reflect area characteristics.

Stay areas and the corresponding stay temporal information need to be encoded into one-hot vectors for learning dataset.
Regarding a stay area, each area is assigned an ID, so it is represented by a one-hot vector with a flag set at the corresponding position in an N-dimensional vector.
For stay temporal information, a vector with a dimension equal to \textit{N'} is prepared, and it is represented by a one-hot vector with a flag set at the corresponding position in an \textit{N'}-dimensional vector.
Pairs of one-hot vectors for a stay area and its stay temporal information are input into Area2Vec for training.

After the completion of training and obtaining the area embeddings, normalization, i.e., conversion to unit vectors, of these vectors is performed.
The reason for this operation is as follows: the euclidean norm of the embeddings obtained from Area2Vec varies in length depending on the amount of data used for training.
In other words, UASs with a large amount of data tends to have a longer norm, while UASs with a small amount of data tends to have a shorter norm. 
Therefore, different amounts of data for areas with similar usage may place them closer together in the vector space on a similarity basis, but farther apart on a distance basis.
To address this issue, normalization is conducted.

Finally, the UASs are subjected to clustering.
The rationale for this operation is as follows.
Interpreting the embedded meaning from each UAS is impossible since it is just a series of numbers, a vector.
Therefore, clustering areas with similar characteristics is necessary to increase abstraction and aggregate information.
By graphing the aggregated information, interpretations can be assigned to UASs belonging to each cluster.
For instance, UASs assigned to Cluster \#1 can be interpreted as "Residential areas," and UASs assigned to Cluster \#2 can be interpreted as "Office areas."
The required number of clusters depends on the desired level of abstraction.
For a coarse examination of urban, fewer clusters may suffice, while a more detailed that may require an increased number of clusters.
In the next section, we start with a small number of clusters and gradually increase the number to confirm the emergence of UASs with various characteristics.
Additionally, by further increasing the number of clusters, we confirm the ability to detect subtle changes in people's behaviors from UASs.

% 4
\section{Evaluation} \label{sec:evaluation}
In this section, we experimentally evaluate whether Area2Vec can capture the characteristics of areas and assert the utility of UAS.

\subsection{Experimental Settings} \label{subsec:experimental_settings}
\subsubsection{Location data} \label{subsubsec:positioning_data}
The location data we use was GPS data collected from apps installed on the user's smartphone with prior consent.
A user's GPS data is a sequence of timestamped coordinates as follows: $\textit{T} = \textit{p}_{0} \rightarrow \textit{p}_{1} \rightarrow \cdots \rightarrow \textit{p}_{k}$, where $\textit{p}_{i} = (\textit{x}, \textit{y}, \textit{t})$, $\textit{i} = 0, 1, \dots, \textit{k}$.
$(\textit{x}, \textit{y}, \textit{t})$ represent latitude, longitude, and timestamp, respectively, where $\textit{p}_{i+1}.\textit{t} > \textit{p}_i.\textit{t}$.

To create UAS, it is necessary to estimate stay information from location data.
This was performed using the method proposed by Iwata et al.~\cite{Iwata2017staypoint}.
Stay places \textit{SP} are represented as a sequence of continuous stay information \textit{I}:
$\textit{SP} = \textit{I}_{0}, \textit{I}_{1}, ..., \textit{I}_{n}$, where $\textit{I} = (\textit{x}, \textit{y}, \textit{d}, \textit{s}, \textit{a})$.
The value of \textit{n} represent the number of stay places and varies among users.
$(\textit{x}, \textit{y}, \textit{d}, \textit{s}, \textit{a})$ denote latitude, longitude, stay time, arrival time, and stay area, respectively, where $\textit{I}_{i+1}.\textit{s} \geqq \textit{I}_{i}.\textit{s} + \textit{I}_{i}.\textit{d}$ with $0 \leqq \textit{i} \leqq \textit{n}$.

\begin{figure}[t]
    \includegraphics[width=\linewidth]{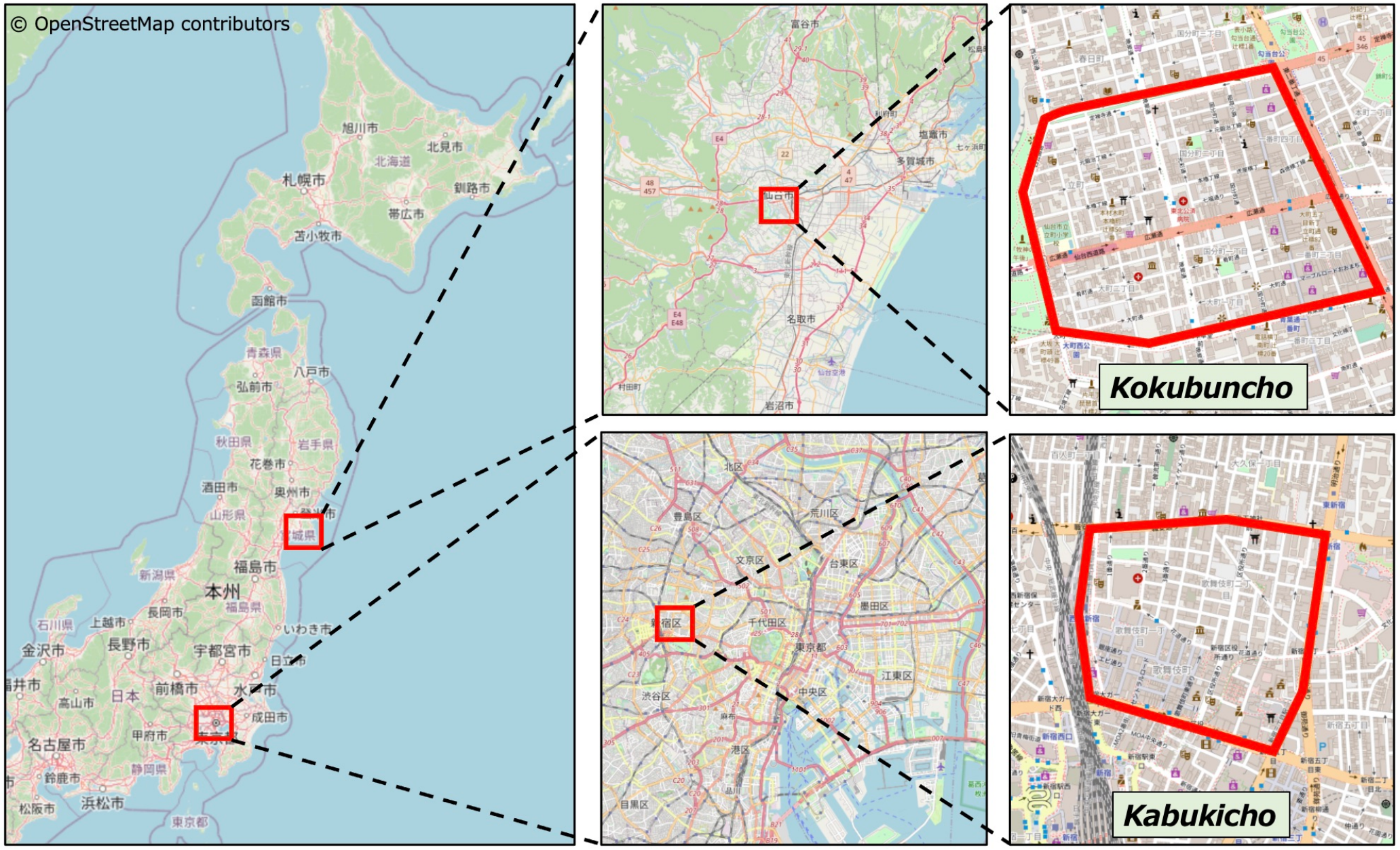}
    \caption{Target districts in the experiment. This includes two districts in Japan: Kabukicho in Shinjuku, Tokyo, and Kokubuncho in Sendai, Miyagi.}
    \label{fig:target_cities}
\end{figure}

\begin{figure*}[t]
    \begin{minipage}[b]{0.6\linewidth}
      \begin{center}
          \includegraphics[width=1\linewidth]{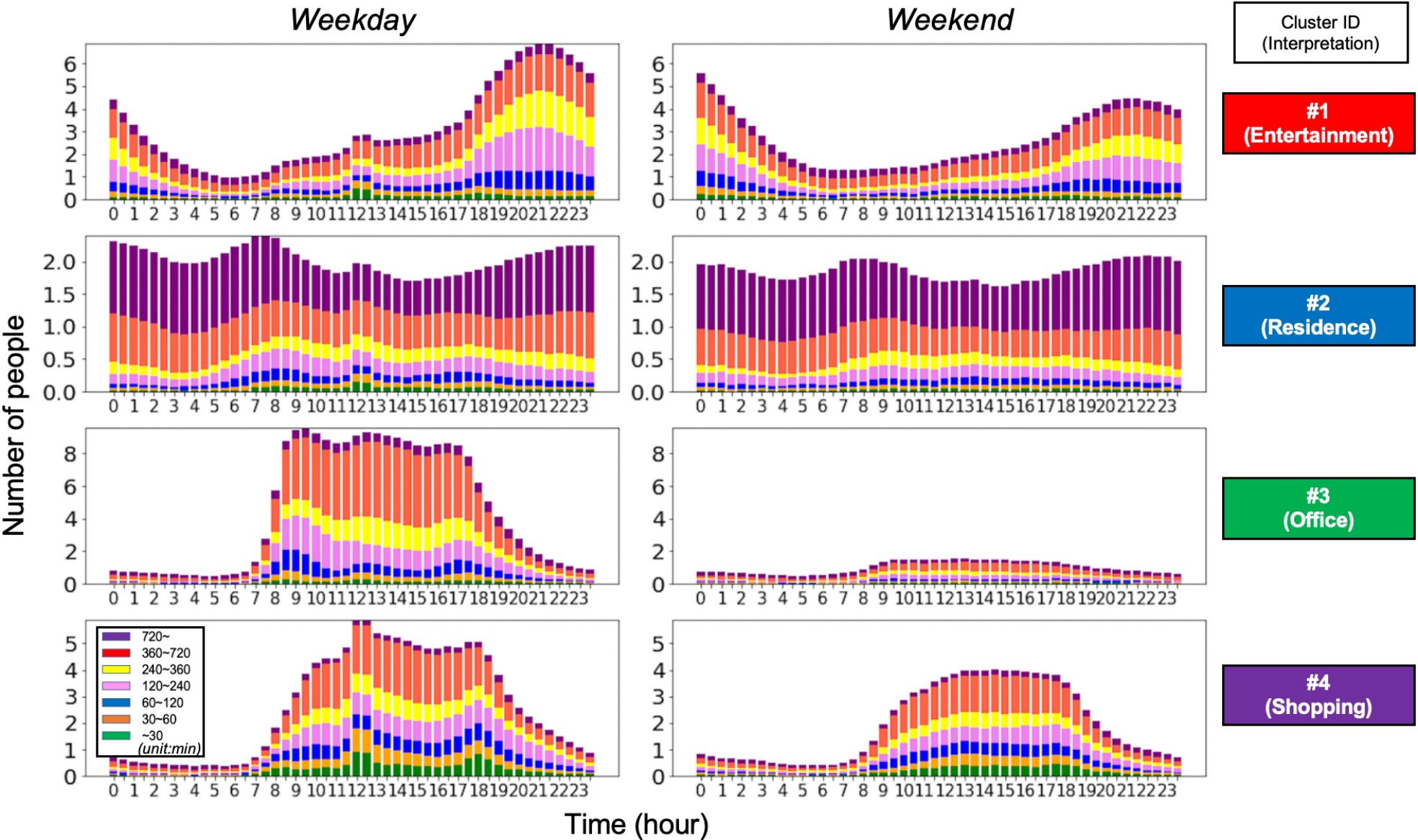}
          \subcaption{Stacked bar graphs for visualizing the characteristics of each cluster, and their interpretation.}
          \label{fig:result_kokubuncho_4cls}
      \end{center}
    \end{minipage}
    \begin{minipage}[b]{0.02\linewidth}
        \begin{center}
        \end{center}
    \end{minipage}
    \begin{minipage}[b]{0.3\linewidth}
        \begin{center}
          \includegraphics[scale=0.27]{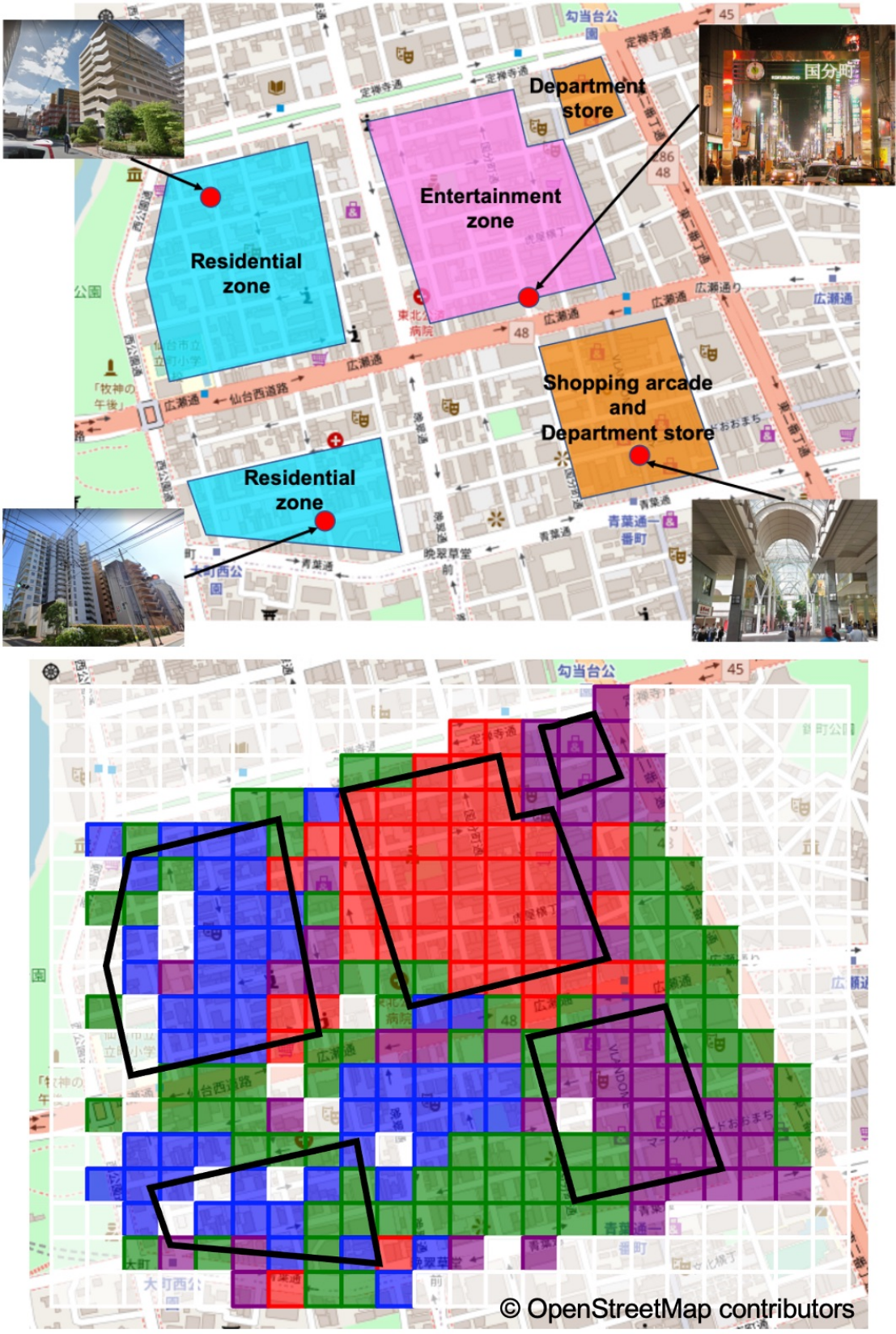}
          \subcaption{Description of the characteristics of each zone of Kokubuncho and reflection of the modeling results on the map.}
          \label{fig:result_kokubuncho_map}
        \end{center}
    \end{minipage}
    \caption{Results of applying our area modeling method to Kokubuncho using the location data from March 2020 and clustering. (a) illustrates the number of people categorized by stay time during each time period, graphed for each cluster. The horizontal axis represents time, and the vertical axis represents the number of people. The colors in each layer represent different stay time. Additionally, interpretations are provided to each cluster based on the characteristics observed from the graphs. (b) illustrates the maps explaining the characteristics of each zone in Kokubuncho and maps reflecting the area modeling result.}
    \label{fig:clustering_result_of_areamodeling}
\end{figure*}

\subsubsection{Target Districts} \label{subsubsec:target_city}
In this paper, we evaluated the effectiveness of Area2Vec from two perspectives:
\begin{inparaenum}[(i)]
    \item capturing the functional characteristics of each area and
    \item capturing area usage changes due to events over time.
\end{inparaenum}
For (\ref{process_1}), we conducted experiments in Kokubuncho, Sendai, Miyagi, Japan (red box above Figure \ref{fig:target_cities}, $900 m \times 1100 m$).
This district encompasses various functionalities, commercial zone (including department stores and shopping arcade), entertainment zone (including bars and clubs), residential zone, and office zone.
The location data used for this experiment were collected in March 2020, comprising 5,181,619 records from 7,348 users, with the estimation of 135,702 stays.

Regarding (\ref{process_2}), we chose Kokubuncho and Kabukicho, Shinjuku, Tokyo, Japan (red box below Figure \ref{fig:target_cities}, approximately $600 m \times 800 m$) as experimental subjects to observe area usage changes influenced by COVID-19.
Kabukicho is a representative "nightlife district" in Japan and is presumed to be significantly affected by COVID-19.
Analyzing this district is expected to provide insights for businesses and government measures against COVID-19.
For Kokubuncho, we used the location data collected from March to July 2020, with 21,137,761 records from 7,348 users, and the estimation identified 538,934 stays.
For Kabukicho, we used the location data collected in April 2019, unaffected by COVID-19, and collected in April 2020 during the state of emergency.
The April 2019 location data consisted of 5,361,112 records with 17,897 users, resulting in 138,082 stays.
The April 2020 location data consisted of 2,482,401 records with 7,734 users, resulting in 57,100 stays.

We used location data from different districts and periods in this paper.
The reason for using the data for the period from March to July 2020 was to compare pre- and post-COVID-19 and to analyze changes in people's behavior over several months.
The reason for using the data for the periods April 2019 and April 2020 was to ensure that the aforementioned changes are not due to seasonality.
And the reason for using different districts for each of the two periods described above was to reinforce the persuasiveness of the insights from the experimental results.

\begin{figure*}[t]
  \begin{minipage}[b]{0.5\linewidth}
    \begin{center}
      \includegraphics[scale=0.45]{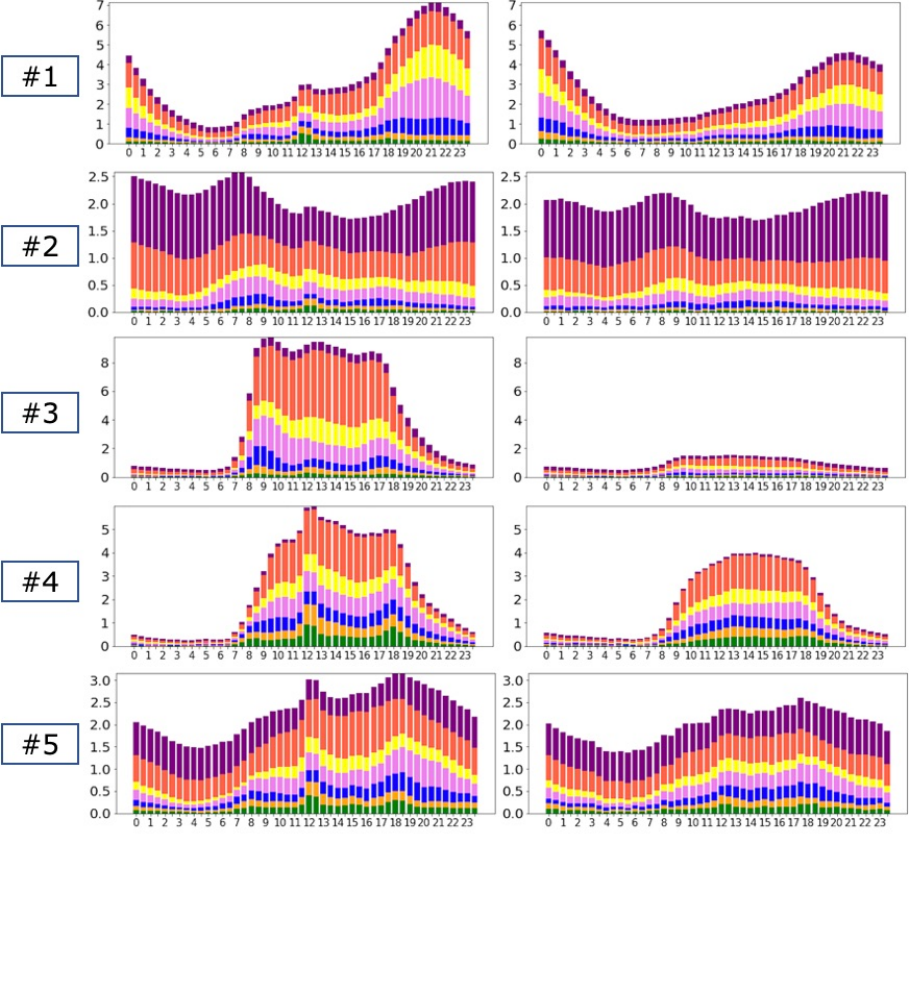}
      \subcaption{Stacked bar graphs with 5 clusters.}
      \label{fig:result_kokubuncho_5cls}
    \end{center}
  \end{minipage}
  \begin{minipage}[b]{0.45\linewidth}
    \begin{center}
      \includegraphics[scale=0.45]{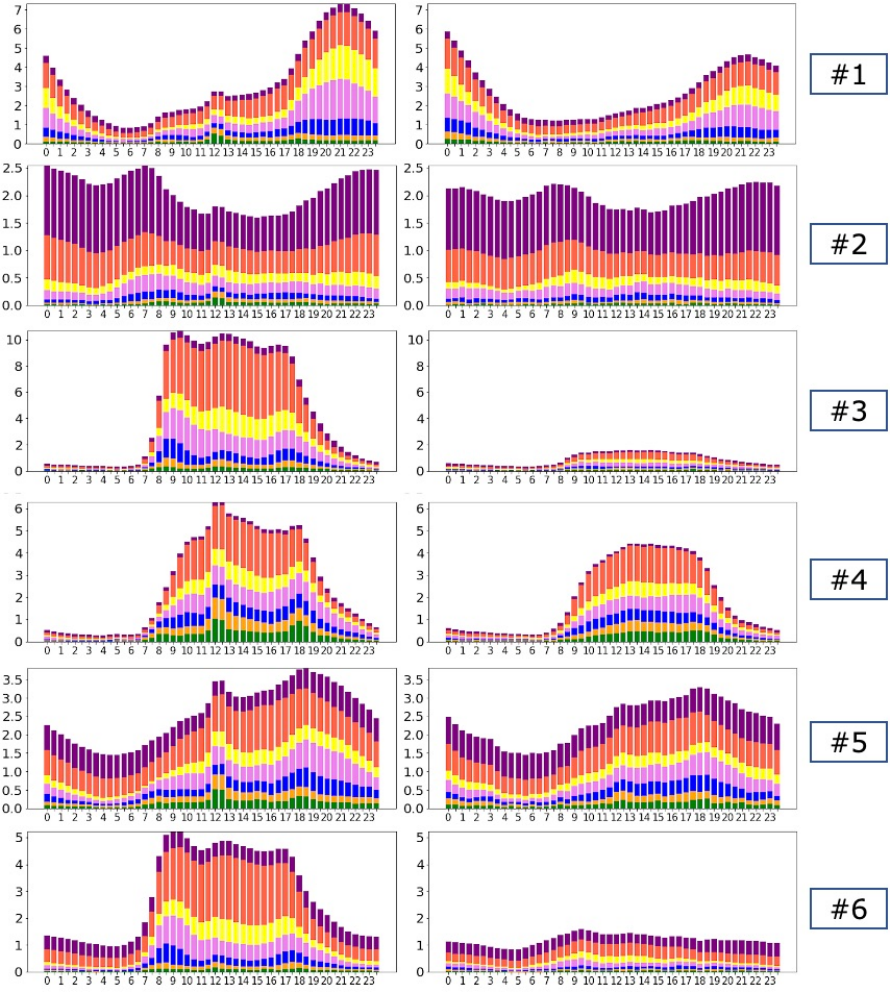}
      \subcaption{Stacked bar graphs with 6 clusters.}
      \label{fig:result_kokubuncho_6cls}
    \end{center}
  \end{minipage}
  \caption{Results of changing the number of clusters in the clustering. This experiment shows that, unlike when the number of clusters is 4, increasing the number of clusters to 5 or 6 allows a more detailed analysis.}
  \label{fig:result_kokubuncho_5and6cls}
\end{figure*}

\begin{figure*}[t]
    \begin{minipage}[b]{0.61\linewidth}
        \begin{center}
            \includegraphics[width=1\linewidth]{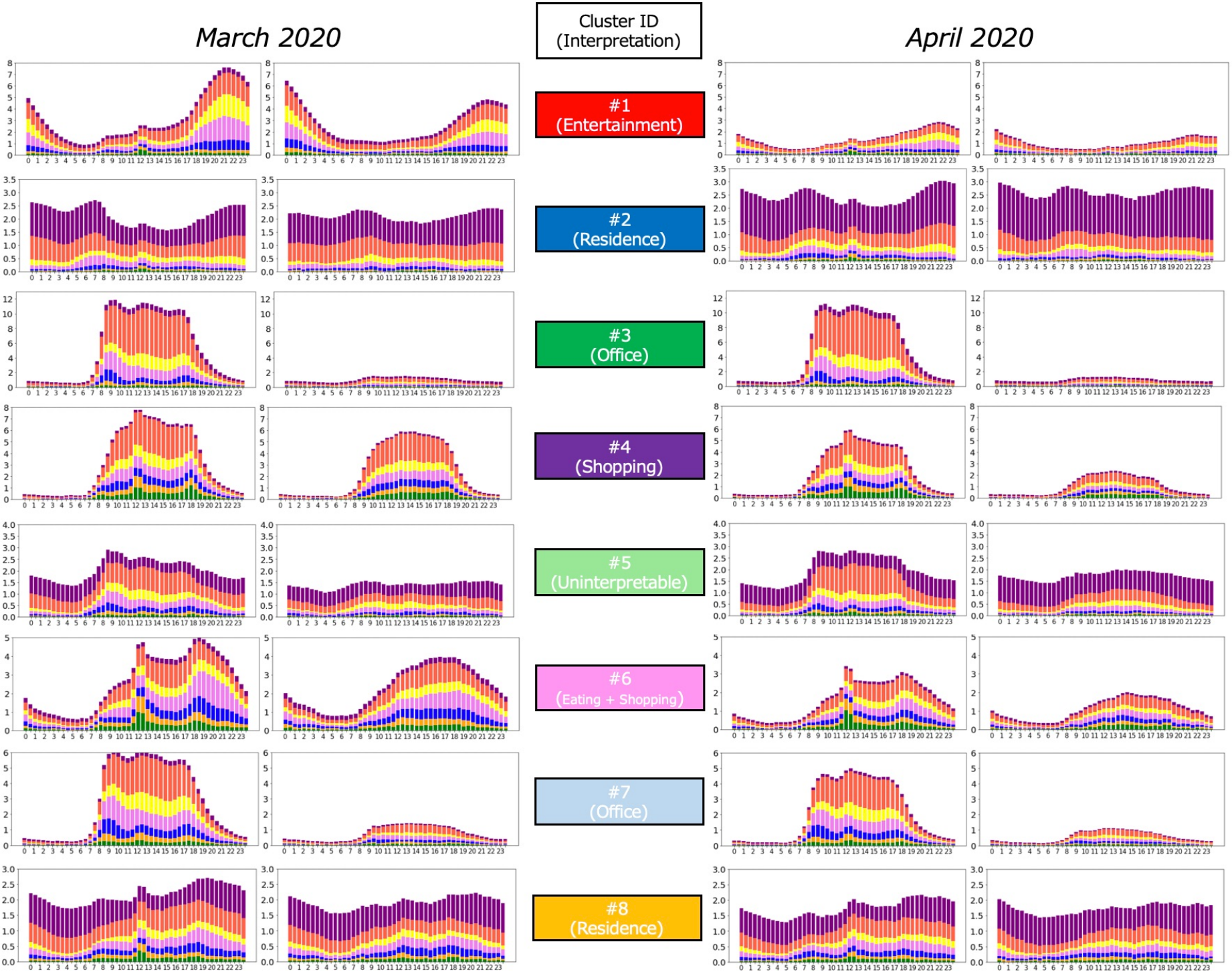}
            \subcaption{Stacked bar graphs of the area modeling results into 8 clusters for the respective periods in March and April 2020.}
            \label{fig:result_kokubuncho_8cls_March_and_April}
        \end{center}
    \end{minipage}
    \begin{minipage}[b]{0.01\linewidth}
        \begin{center}
        \end{center}
    \end{minipage}
    \begin{minipage}[b]{0.3\linewidth}
        \begin{center}
            \includegraphics[width=1\linewidth]{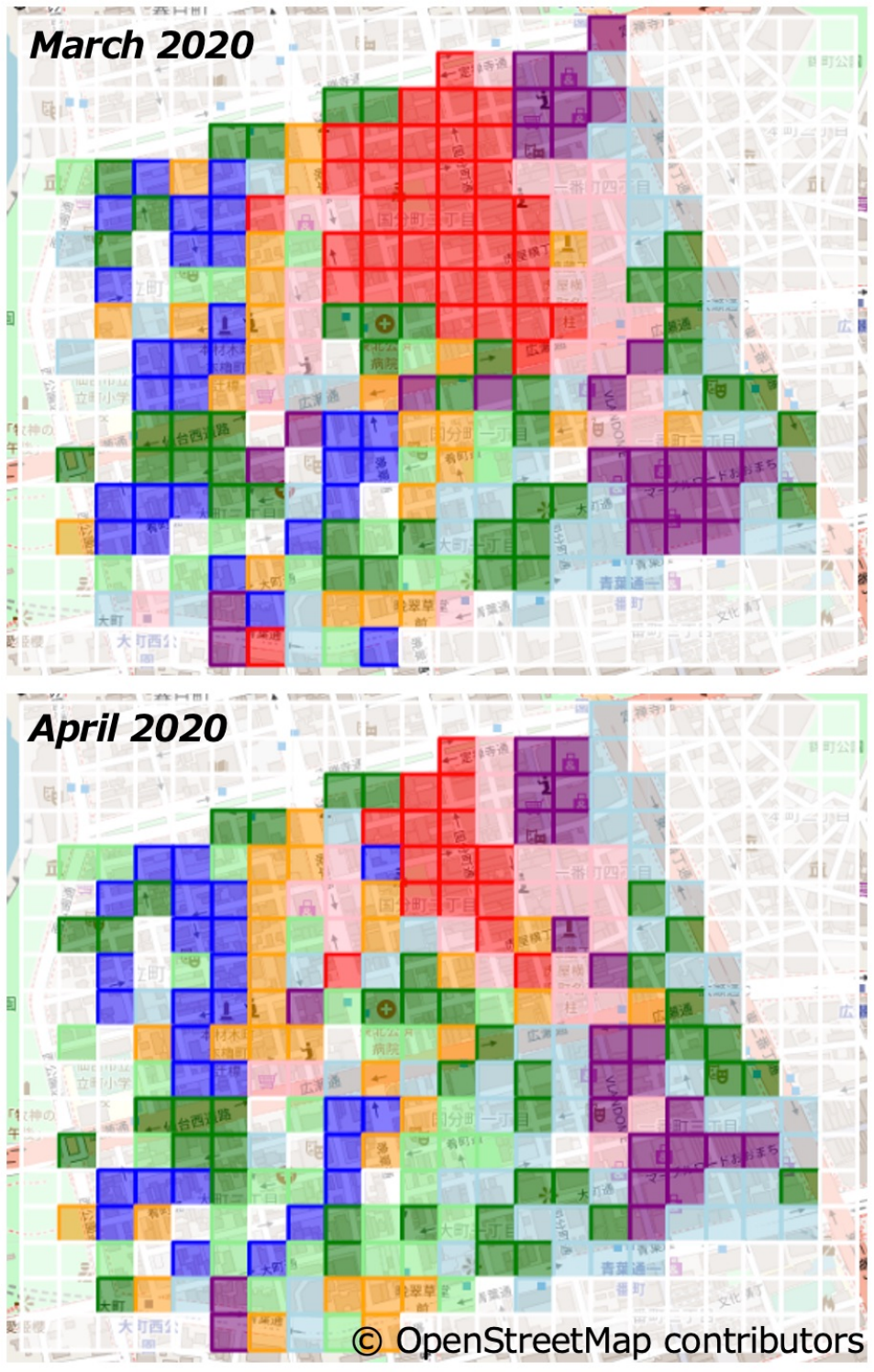}
            \subcaption{Maps reflecting the area modeling results classified into 8 clusters for the respective periods in March and April 2020.}
            \label{fig:result_kokubuncho_map_March_and_April}
        \end{center}
    \end{minipage}
    \caption{Comparison of area usage changes for Kokubuncho during the unaffected and affected periods of COVID-19. This experiment aims to observe area usage changes influenced by COVID-19 by comparing March and April of 2020. March represents a period when there were no restrictions on activities due to COVID-19, while April corresponds to a period when people had refrained from going out due to the declaration of a state of emergency. Observing (a) and (b), it is evident that there was a decrease in human traffic to entertainment areas at night and a shift in people's lifestyles from night-oriented to day-oriented.}
    \label{fig:result_change_march_and_april}
\end{figure*}

\subsection{Analysis of UAS} \label{subsec:UAS_analysis}
As described in \ref{subsubsec:target_city}, area modeling is performed for Kokubuncho to see if it is consistent with human senses in the classification of urban functions.

We meshed this district and generated a distributed representation of grids.
The grid size was set to 50 meters on each side with the intention of mitigating the impact of GPS data positioning errors.
Although the total number of grids was initially 396, we reduced it to 241 by excluding grids with fewer than 100 recorded stays.
This restriction serves to decrease computational complexity and prevent biased characterizations by a small number of users.
Subsequently, we refer to each grid as an "area," and we describe how to analyze the UASs of the 241 areas created by Area2Vec.

Following the completion of making UASs, the next step involves clustering.
We employed k-means++ algorithm for clustering.
Clustering creates semantic cohesion and facilitates the interpretation of UASs.
The clustering results for the UASs of 241 areas are depicted in Figure \ref{fig:clustering_result_of_areamodeling}.
The stacked bar graph in Figure \ref{fig:result_kokubuncho_4cls} illustrates the distribution of the number of people based on the length of stay time in each cluster.
The left and right columns correspond to weekdays and weekends including holidays, respectively.
The horizontal axis represents time in 30-minute bins, and the vertical axis denotes the number of people.
Notes that the vertical axis is normalized by dividing by the number of UASs assigned to each cluster, since the number of UASs assigned to each cluster differs.
Additionally, given the discrepancy in the number of weekdays and weekends within the period, normalization is performed based on the respective day counts.
In other words, this graph illustrates the distribution of the number of people based on the length of stay time per area and per day.
Through this process, comparisons between clusters and between weekdays and weekends become feasible.
Also note that those who have prolonged stays are counted in multiple bins.
For instance, if someone stayed from 10:00 am to 12:00 pm, he would be counted in all bins from the 20th to the 24th.
This graphical representation enables the interpretation of the characteristics of UASs within each cluster.
Subsequently, we proceed with interpreting the UASs assigned to each cluster.

We set the number of clusters to 4.
The rationale behind choosing 4 clusters is, as discussed in \ref{subsubsec:target_city}, to broadly categorize the functions of this district into 4 types: shopping, entertainment, residential, and office zones.

\begin{description}
    \item[\textit{Cluster \#1}]: This cluster can be interpreted as "entertainment area," consisting of establishment such as bars and clubs, for the following reasons.
      \begin{itemize}
        \item Discrepancy in the number of people during day and night.
        \item Medium-term (between 120 and 360 minutes) stays peaking around 21:00 during the night.
      \end{itemize}
      
    \item[\textit{Cluster \#2}]: This cluster can be interpreted as "residential area," characterized primarily by homemakers, for the following reasons.
      \begin{itemize}
        \item Stays exceeding 12 hours throughout both weekdays and weekends.
        \item Long-term (over 360 minutes) stays peaking around 21:00 during the night.
      \end{itemize}
    
    \item[\textit{Cluster \#3}]: This cluster can be interpreted as "office area," primarily used by office workers, for the following reasons.
      \begin{itemize}
        \item Long-term stays starting around 8:00 on weekdays.
        \item Low number of people during the night and on weekends.
      \end{itemize}

    \item[\textit{Cluster \#4}]: This cluster can be interpreted as "shopping area," including eating establishments, for the following reasons.
      \begin{itemize}
        \item Large number of short-term (120 minutes or less) stays during daytime.
        \item Large number of people even on weekends.
        \item Increase in the number of people at mealtimes.
      \end{itemize}
\end{description}

\begin{figure*}[t]
    \begin{minipage}[b]{0.95\linewidth}
        \centering
        \includegraphics[scale=0.4]{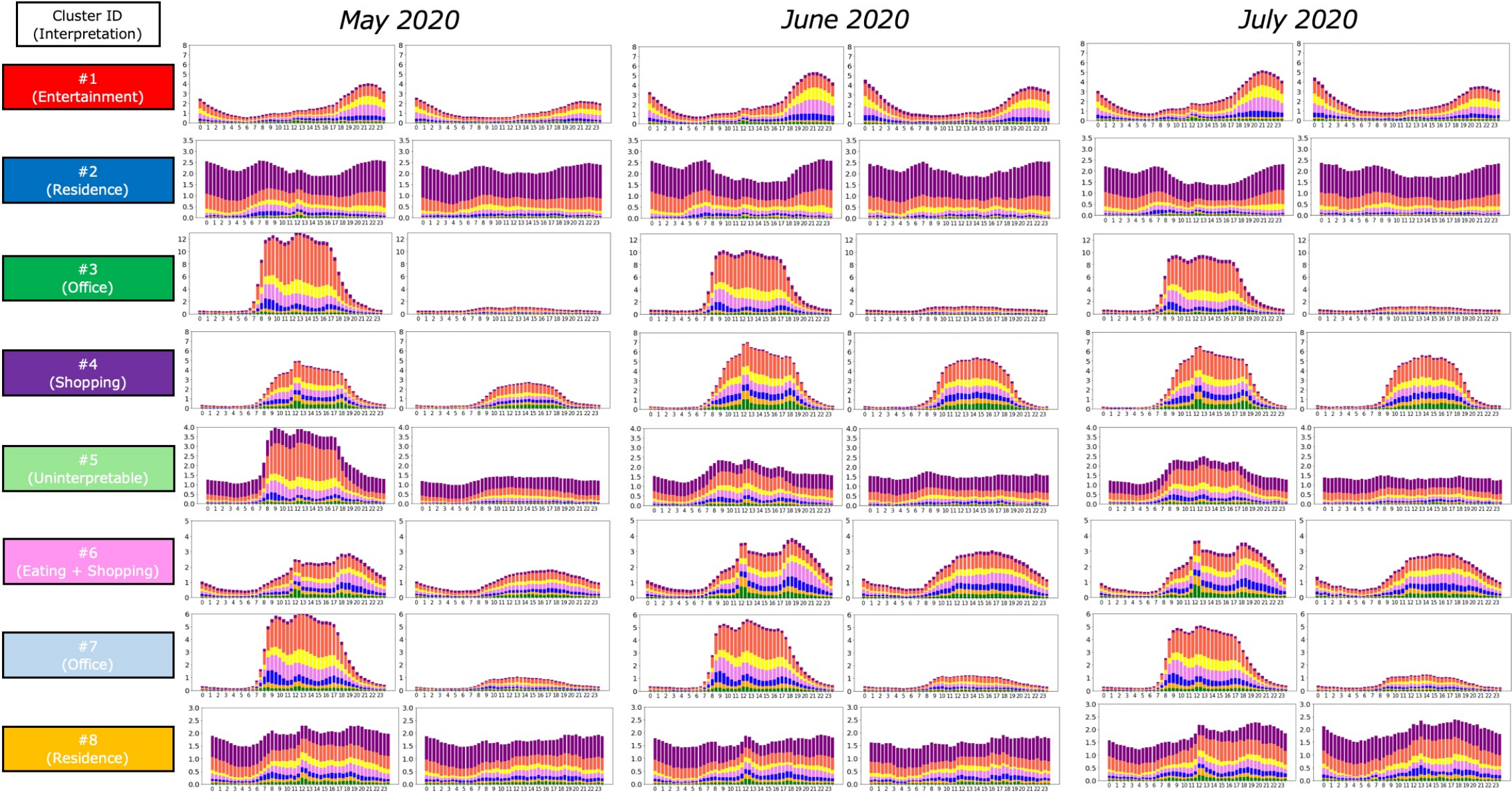}
        \subcaption{Stacked bar graphs of the area modeling results classified into 8 clusters for each period from May to July 2020.}
        \label{fig:result_kokubuncho_8cls_May_to_July}
    \end{minipage}\\
    \begin{minipage}[b]{0.95\linewidth}
        \centering
        \includegraphics[scale=0.4]{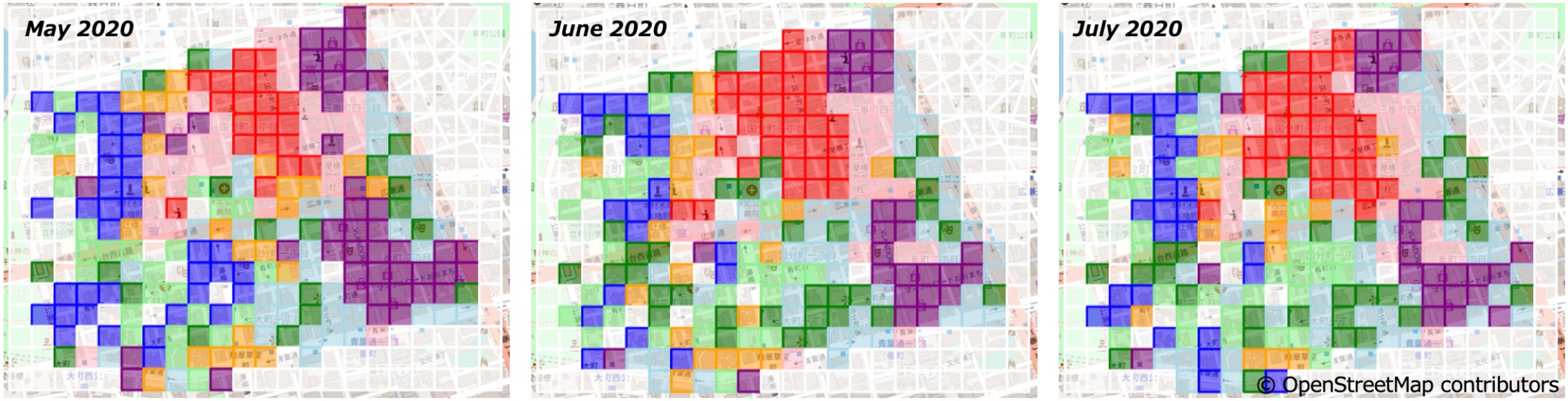}
        \subcaption{Maps reflecting the area modeling results classified into 8 clusters for each period from May to July 2020.}
        \label{fig:result_kokubuncho_map_May_to_July}
    \end{minipage}
    \caption{Area usage changes in Kokubuncho over several months due to the impact of COVID-19. Observing (a) and (b), it can be anticipated that it would take a bit more time for human traffic in this area to fully return to normal. On the other hand, it becomes apparent that there were individuals who reverted to their original lifestyle, suggesting a weakening of the sense of crisis regarding COVID-19.}
    \label{fig:result_change_may_to_july}
\end{figure*}

Figure \ref{fig:result_kokubuncho_map} shows the clustering results drawn on a map.
The left map provides a classification of Kokubuncho's functions, while the right map color-codes each area based on the assigned cluster.
Upon comparing these two maps, it is evident that the area classification aligns roughly with our expectations.
Areas characterized by entertainment areas are marked in red (\#1), residential areas in blue (\#2), and shopping arcade and department stores in purple (\#4).
Along major roads where office buildings are scattered, these areas are designated in green (\#3).
The white areas represent areas excluded from the analysis due to having fewer than 100 stays, as discussed in \ref{subsec:UAS_analysis}.
Consequently, it is apparent from these observations that the UASs reflect the distinctive characteristics of the 4 functions, i.e., shopping, entertainment, residence, and office in this district.

One of the strengths of UAS is the ability to change the granularity of the analysis by changing the number of clusters.
Hence, we explored the differences when changing the cluster number.
Figure \ref{fig:result_kokubuncho_5and6cls} presents stacked bar graphs for each cluster when the cluster numbers are set to 5 and 6.
First, in the case with 5 clusters (Figure \ref{fig:result_kokubuncho_5cls}), clusters \#1–\#4 can be given the same interpretations as 4 clusters, while cluster \#5 portrays a new area characteristic.
It exhibits a distinctive feature of high foot traffic even past midnight and increased activity during mealtimes on weekdays, along with long-term stays throughout the day.
This suggests that the UASs in this cluster reflect the characteristics of the entertainment area and the eating establishments and residences located facing it.
Turning to the case with 6 clusters (Figure \ref{fig:result_kokubuncho_6cls}), clusters \#1–\#5 can be given the same interpretations as 5 clusters, but cluster \#6 unveils a new area characteristic.
It is similar to \#3, but stands out with more frequent stays exceeding 720 minutes and a higher proportion of stays occurring on weekends.
Thus, the UASs assigned here are a kind of "office" characterized by frequent occurrences of long-term stay and occasional work on holidays.
This demonstrates that increasing the cluster number enables a more detailed analysis.

\begin{figure*}[t]
    \centering
    \includegraphics[scale=0.5]{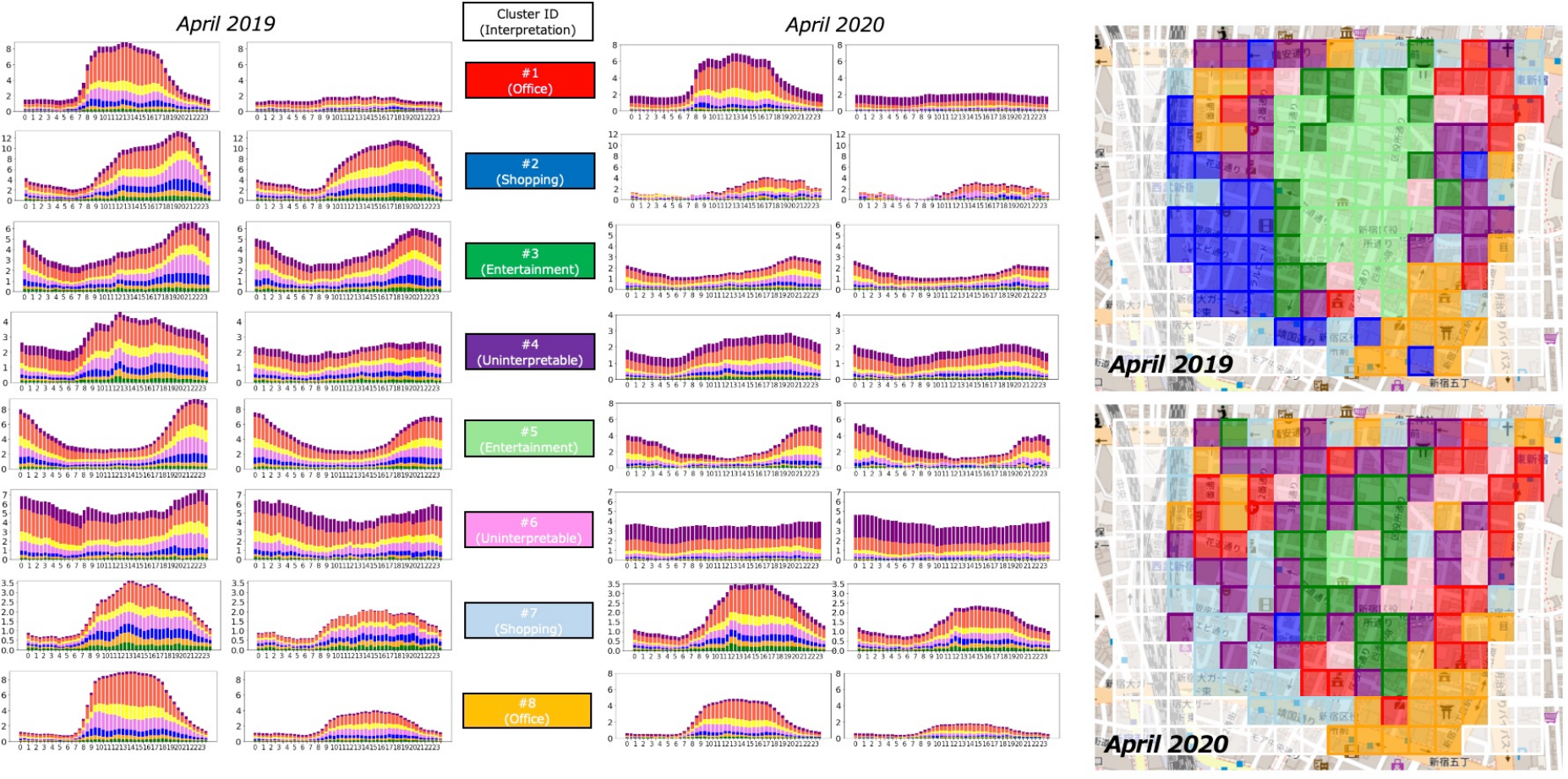}
    \caption{Results of applying our area modeling method to Kabukicho using the location data from April 2019 and 2020 and clustering. This experiment supports the claim that area usage changes observed by the experiments conducted so far are not seasonal, but are the result of COVID-19.}
    \label{fig:result_kabukicho_8cls_map}
\end{figure*}

\subsection{Area usage changes due to COVID-19 impact}
We analyzed area usage changes to understand how people's behavior were affected by COVID-19. 
We conducted the analysis in Kokubuncho and Kabukicho.

\subsubsection{Kokubuncho} \label{subsubsection:covid19_kokubuncho}
Similar to the procedures outlined in \ref{subsec:UAS_analysis}, we perform meshing of the target district, create UASs, and apply clustering.
The resulting characteristics of each cluster are depicted in Figure \ref{fig:result_kokubuncho_8cls_March_and_April}.
We initially compared the months of March and April.
Unlike \ref{subsec:UAS_analysis}, we set the number of clusters to 8 to capture finer changes in people's behavior.
The clustering results are color-coded and drawn on the map as Figure \ref{fig:result_kokubuncho_map_March_and_April}.

The observed changes in people's behavior from these figures are as follows.
The stacked bar graphs (Figure \ref{fig:result_kokubuncho_8cls_March_and_April}) indicate a significant decrease in the number of people during weekends in cluster \#4 (shopping area).
This suggests that people refrained from non-essential outings during weekends.
In addition, the number of people in cluster \#1 (entertainment area) and \#6 (eating establishment and shopping area) decrease significantly regardless of weekdays or weekends, and this also indicates that people are forced to refrain from non-essential outings.
Next, the number of people in cluster \#7 (office area) also decreased.
This indicates that the number of people coming to work decreased.
We could possibly say that the number of people who work remotely increased.
Lastly, in cluster \#2 (residential area), unlike other clusters, there is an increase in the number of people, suggesting that people spent more time at home in response to the call for refraining from going out.

We next conducted an analysis using the map reflecting the clustering results (Figure \ref{fig:result_kokubuncho_map_March_and_April}).
It is evident that the red areas (\#1, entertainment area) significantly decreased.
This indicates that the number of people who come to the entertainment zone at night decreased due to the request for refraining from non-essential outings.
The next step is to find out how the area usage that was entertainment area in March changed in April.
The map provides insights into this transformation.
First, the western and southern parts of the entertainment areas were replaced by light blue (\#7, office area) and orange (\#8, residential area).
This suggests that these areas are predominantly occupied by office workers or residents, reflecting their behavioral characteristics in the UASs.
Then, the eastern part of the entertainment areas were replaced by pink (\#6, eating establishment and shopping area), implying a shift in the behavior of people in these areas from nightlife to daytime activities.

Figure \ref{fig:result_change_may_to_july} presents stacked bar graphs created from the data for each month from May to July, along with clustering results reflected in the maps.
Examining the values on the vertical axis of Figure \ref{fig:result_kokubuncho_8cls_May_to_July}, it is apparent that human traffic into the entertainment areas and shopping areas was gradually recovering, but a complete return to pre-COVID-19 levels is expected to take some more time.
However, when observing Figure \ref{fig:result_kokubuncho_map_May_to_July}, it is clear that the areas were returning to their previous area usage.
In other words, while these visualizations reflecting clustering results indicate that some individuals are still practicing self-restraint, they simultaneously show that others have returned to their normal lifestyle.

However, we can not deny the possibility that the changes in people's behavior described so far are seasonal in nature.
The next step involves investigating the extent of the impact of COVID-19 by comparing data for the same season in April 2019 and April 2020 in Kabukicho.

\subsubsection{Kabukicho} \label{subsubsection:covid19_kabukicho}
Figure \ref{fig:result_kabukicho_8cls_map} depicts the results for Kabukicho in April 2019 and April 2020, illustrating them through stacked bar graphs and maps.
The number of clusters was set to 8, consistent with \ref{subsubsection:covid19_kokubuncho}.
The observed changes in people's behavior inferred from these figures are as follows.
The first thing that can be read from the graphs is that stays of 720 minutes or longer (purple layer) are noticeable throughout.
What this means is that the proportion of people with long-term stay increased more than those with short- to medium-term stay.
In other words, it can be inferred that customers decreased, and the behavior of employees became majority.
Another noteworthy observation is the decrease in the number of people across all clusters.
This implies an overall reduction in human traffic into Kabukicho.
Since the data amount is completely different between 2019 and 2020, it is possible to predict how human traffic changed from that information, but Area2Vec allows for intuitive analysis through visualization.

Examining the maps reflecting clustering results, it is evident that the green and lime green areas (\#3 and \#5) decreased, transitioning to purple (\#4).
What this means is a reduction in nighttime usage to the extent that people's daytime behavior was reflected.
There are also areas changing to pink (\#6).
This is considered to be because the characteristics of those who reside or operate businesses there were reflected more than those who came as customers.
Focusing on specific facilities, "Toho Building (Cinema, \textit{A} on the map)" transitioned from blue (\#2) to light blue (\#7).
This implies that the main time period of use shifted from the evening to the daytime.
As for the "Ward Office (\textit{B} on the map)," it remains unchanged, retaining red (\#1).
This suggests that, even in the COVID-19 pandemic, the ward office continued operations as it couldn't stop working.

This subsection examined the differences between April 2019 and April 2020 in Kabukicho.
Despite the same season, significant changes in people's behavior are evident.
In \ref{subsubsection:covid19_kokubuncho}, we observed changes in people's behavior over several months in Kokubuncho, and we can conclude that these changes were not seasonal, but were largely influenced by COVID-19.

In this section, using Area2Vec, we examined the changes in people's behavior due to the impact of COVID-19 in two districts, Kokubuncho and Kabukicho.
The analysis reveals the following findings.
First, human traffic was greatly reduced by COVID-19.
Second, the overall behavior of people shifted to the daytime.
This suggests that, within the context of COVID-19, individuals were refraining from non-essential outings such as visiting entertainment zone at night.
Additionally, the fact that human traffic did not fully recovered over the months suggests that some people were still refraining, while a certain number of people were returning to their previous lifestyles.
This indicates that there was a gap in the sense of urgency for COVID-19 among individuals.

% 5
\section{Summary} \label{sec:summary}
We proposed an area modeling method named Area2Vec, which characterizes areas based on their usage.
This model is inspired by the Word2Vec and creates a distributed representation of areas.
For characterizing areas, we use three types of stay temporal information: day of the week, arrival time, and stay time.
These information alone allow area modeling to account for the effects of temporal events such as urban development, seasonal changes, and pandemics.

In experiment, Area2Vec was evaluated by classifying the functions of areas and by investigating area usage changes due to the impact of COVID-19.
For the former, we estimated the function of each area from the modeling results in a district that can be divided into four functional categories, i.e., shopping, office, residential, and entertainment zone, and show the usable of Area2Vec from the certainty of the estimated results.
For the latter, we observed a decrease in human traffic and a shift of people's behavior from night to day in a entertainment zone from the area modeling results, and demonstrate the usefulness of Area2Vec.
We were also able to capture how people's lives were returning to normal with each passing month, and we could see that they were tired of refraining from going out and that their sense of urgency about COVID-19 was waning.

In the future work, the following two approaches need to be considered.
The first is to characterize areas considering the sequential relationship of movements.
There are some regularities in human behavior, such as moving from home to office in the morning and vice versa in the evening.
While this information could be valuable for characterizing areas, it has not been accounted for in Area2Vec.
Incorporating this aspect could lead to a more detailed area modeling approach.
The second is to involve using the results of Area2Vec to represent people's trajectory.
In existing studies on modeling people's trajectories, data represented by POI and semantics information, which are difficult to collect, are used.
By replacing this with a vector that reflects the area usage, it becomes possible to analyze people's movements without collecting any data other than location data, which will lead to further utilization of location data.

\section*{Acknowledgment}
This research was supported in part by JST CREST (JPMJCR1882), the Ministry of Internal Affairs and Communications SCOPE, and NICT commissioned research (222C0101). 
We would like to express our gratitude to Blogwatcher, Inc.\footnote{https://www.blogwatcher.co.jp/} for their cooperation in providing the data.

\bibliographystyle{ACM-Reference-Format}
\bibliography{ref}

\end{document}